\documentclass[conference]{IEEEtran}
\IEEEoverridecommandlockouts
\usepackage{cite}
\usepackage{amsmath,amssymb,amsfonts}
\usepackage{algorithmic}
\usepackage{graphicx}
\usepackage{textcomp}
\usepackage{xcolor}
\def\BibTeX{{\rm B\kern-.05em{\sc i\kern-.025em b}\kern-.08em
    T\kern-.1667em\lower.7ex\hbox{E}\kern-.125emX}}
\begin{document}

\title{Consistent Video Colorization via Palette Guidance} 

\author{
Han Wang, Yuang Zhang, Yuhong Zhang, Lingxiao Lu, Li Song
}
\author{
    \IEEEauthorblockN{Han Wang, Yuang Zhang, Yuhong Zhang, Lingxiao Lu, Li Song}
    \IEEEauthorblockA{Shanghai Jiao Tong University}
    \IEEEauthorblockA{\{esmuellert, zyayoung, rainbowow, lulingxiao, song\_li\}@sjtu.edu.cn}
}

\maketitle
\begin{abstract}

Colorization is a traditional computer vision task and it plays an important role in many time-consuming tasks, such as old film restoration. Existing methods suffer from unsaturated color and temporally inconsistency. In this paper, we propose a novel pipeline to overcome the challenges. We regard the colorization task as a generative task and introduce Stable Video Diffusion (SVD) as our base model. We design a palette-based color guider to assist the model in generating vivid and consistent colors. The color context introduced by the palette not only provides guidance for color generation, but also enhances the stability of the generated colors through a unified color context across multiple sequences. Experiments demonstrate that the proposed method can provide vivid and stable colors for videos, surpassing previous methods.

\end{abstract}
\section{Introduction}
Video colorization is essential for enhancing the visual experience of historical video materials and old films. This task involves transforming grayscale video sequences into vivid, full-color versions while maintaining temporal consistency across frames. 

However, existing methods still face two primary challenges: \textbf{unsaturated colors} and \textbf{temporal inconsistency}. 
On one hand, the issue of \textbf{unsaturated colors} reflects a common challenge in image colorization, where colors often appear dull or lack diversity. To address this, researchers have successfully integrated generative models~\cite{vitoria2020chromagan, CT2, kumar2021colorization, kim2022bigcolor} and multi-modal priors~\cite{huang2022unicolor, weng2024cad} into image colorization methods, achieving significant improvements in color vividness. 
For these methods, maintaining consistent colors across frames conflicts with the fact that each frame is colored independently.
On the other hand, video colorization introduces the additional challenge of \textbf{temporal inconsistency}—the need to maintain consistent colors across frames. To tackle this, previous work~\cite{Lei_2019_CVPR, zhou2023vcgan, Liu2024Temporally} has employed techniques such as optical flow\cite{ilg2017flownet, sun2018pwc}. While these methods process the video frame by frame, they struggle to achieve long-range coherence and are prone to cumulative errors, which further exacerbate the issue of temporal inconsistency. 
These limitations highlight the need for a more integrated approach that can balance color richness with temporal coherence, ensuring high visual quality in video colorization.


\begin{figure}[t]
\centering
\includegraphics[width=\linewidth]{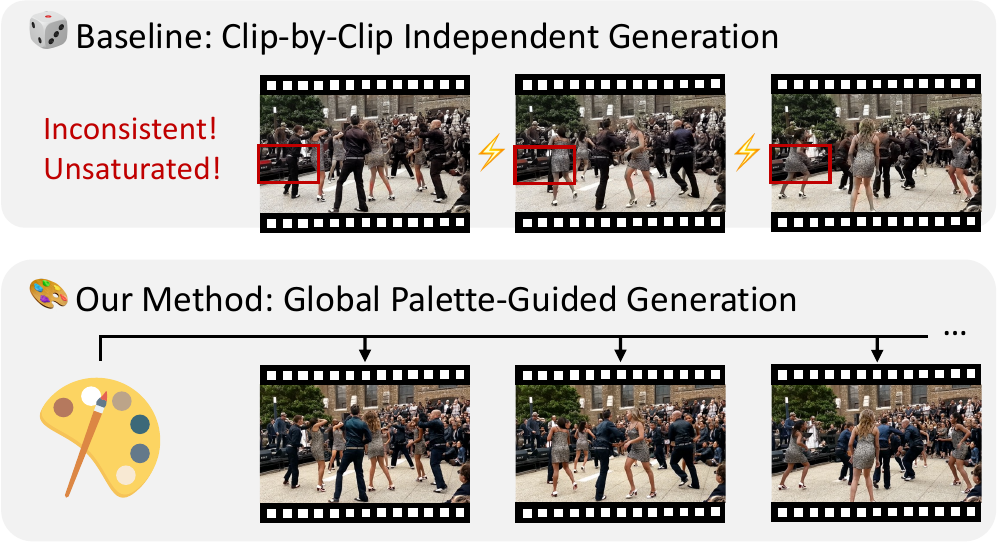}
\caption{Comparison with frame independent colorization framework TCVC\cite{Liu2024Temporally}. Our method shows superiority in color vividness and temporal consistence.}
\label{fig:motivation}
\end{figure}

The rapid development of large-scale generative models has significantly advanced downstream vision tasks\cite{wang2022zero,brooks2023instructpix2pix,xia2023diffir}, including colorization\cite{weng2024cad,liang2024control,cong2024automatic,du2024multicolor,liu2023video, li2024towards, bozic2024versatile}.
Among these, diffusion-based methods have become a cornerstone, using image-to-video diffusion priors to generate semantic reasonable colors. 
Recent studies, such as those in~\cite{liu2023video, li2024towards, huang2024lvcd,bozic2024versatile}, have shown the potential of diffusion model (DM)-based approaches for video colorization. These methods address temporal consistency by integrating optical flow priors or cross-frame attention mechanisms into the image colorization backbone, enabling multi-frame colorization.
For instance, methods like \cite{huang2024lvcd,bozic2024versatile} have introduced temporally deformable attention and cross-clip fusion to maintain long-term color consistency and prevent flickering or color shifts. 
However, these cross-frame attention mechanisms typically refer to only a limited number of adjacent frames, which may not be sufficient for ensuring long-range temporal consistency across the entire video. 
This limitation underscores the need for more robust and comprehensive approaches that can effectively address both color richness and temporal coherence in video colorization.

To address the challenges of video colorization, we propose a color palette-guided video diffusion framework, which enhances both color richness and temporal consistency by leveraging palette for global guidance and allowing for a diverse range of inputs for the palette. 
Our method is based on fine-tuning an image-to-video diffusion model. However, direct fine-tuning often results in generated frames that appear unsaturated and muted. We attribute this issue to two primary factors: (1) the model may fall into a conservative "shortcut" of merely recovering grayscale values, and (2) the output is highly sensitive to the color distribution of the training data, making it prone to biases. 

To address these issues and obtain more saturated colors, we introduce a palette guidance mechanism. The color palette provides a rich color context that can significantly enhance the color saturation of the generated videos.
Furthermore, we identify that temporal consistency across denoising windows remains a significant challenge. Since previous models typically process only a limited number of frames simultaneously, maintaining globally consistent colors throughout longer video sequences becomes difficult. To address this, we utilize the palette functions as a global guide to ensure consistency across frames. Specifically, we process the global palette through a linear layer to obtain color embeddings that guide the denoising process. 
Besides, we found that simply using colors extracted from a single image as the palette condition may not always be suitable. This approach can sometimes lead to results that do not match reality. Our method, however, naturally allows for a wider variety of inputs to serve as the palette guidance. Therefore, we simply introduce more types of inputs. These include randomly sampling colors from a trained Mixture Model(GMM)\cite{reynolds2009gaussian} and using large language models like GPT\cite{achiam2023gpt} to extract colors based on the objects in the image. This helps to better guide the color generation of the image. In summary, the palette can effectively accommodate various types of inputs, converting different forms of colors into a unified domain condition.

Quantitative and qualitative experimental results indicate that our proposed automatic video colorization method outperforms the baseline methods in terms of color saturation and video quality. The innovations of our method are reflected in the following aspects:

1. We develope an integrated diffusion-based framework capable of simultaneously addressing color unsaturation and inter-frame color discontinuity issues.

2. We propose employing a palette as global guidance to resolve long-range instability problems in videos.

3. Our palette naturally accommodates various input formats, providing more solutions for color appropriateness.

\begin{figure*}[t]
\centering
\includegraphics[width=\textwidth]{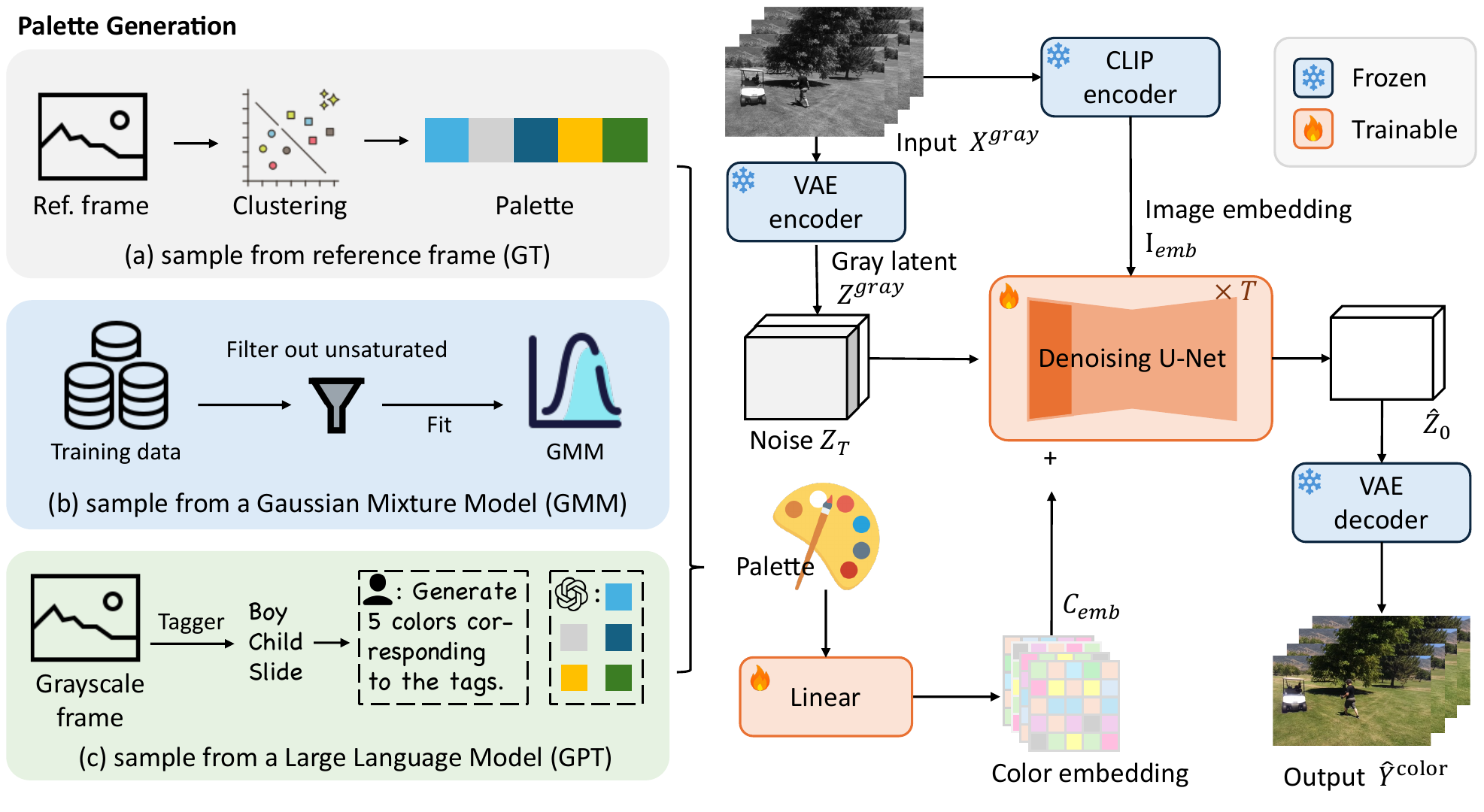}
\caption{Overview of the video colorization pipeline with palette-guidance. We re-purpose a pre-trained video diffusion model and augment it with palette guidance. The left side illustrates three methods for generating palettes. During training, the palette is generated in the manner depicted in (a), with the reference image being a randomly selected frame from each training clip.}
\label{fig:overview}
\end{figure*}

\section{Related Works}
\subsection{Image Colorization}
Methods based on generative models, as opposed to those based on convolutional neural networks, offer richer colors and have thus become the predominant research direction in automatic image colorization. Both Generative Adversarial Networks (GANs)~\cite{vitoria2020chromagan,Jin2022Reproducibility,kim2022bigcolor} and Transformers~\cite{kumar2021colorization,CT2, kang2023ddcolor, du2024multicolor} have made remarkable progress in this task. More recently, Latent Diffusion Models (LDMs)~\cite{rombach2022high}, as a superior alternative to GANs, have begun to attract attention in multi-modal image colorization~\cite{weng2024cad,cong2024automatic,liang2024control}. 
\subsection{Video Colorization}
Existing video colorization methods fall into two categories: 1) exemplar-based video colorization and 2) automatic video colorization. 

\textbf{Exemplar-based video colorization} methods rely on a color exemplar image and propagate the colors to the video frames. Early methods~\cite{zhang2019deep,iizuka2019deepremaster,TRAN2025125437,chen2024exemplar,yang2025colormnet} adopt networks to find the correspondence of gray frames and exemplar image in the deep feature domain. Then the colors are aligned according to the correspondence. Yang et al.~\cite{yang2024bistnet} proposed a bidirectional exemplar-based video colorization method to better propagate the reference colors and avoid the inaccurate matches. This category of methods requires the user to provide reference images that are highly relevant to the video content. In practical applications, frames from the video subsequent to the image colorization process may serve as exemplar. Nevertheless, methods relying on match propagation exhibit insufficient flexibility when confronted with the appearance of new objects.

\textbf{Automatic video colorization} methods operate independently of additional references and employ specialized modules to preserve consistency between frames. Previous GAN-based methods~\cite{antic_deoldify_nodate,Lei_2019_CVPR,zhou2023vcgan,Liu2024Temporally}, similar to the example-based approach, utilize optical flow~\cite{ilg2017flownet,sun2018pwc} to propagate motion information across frames. Even though these methods refer to inter-frame information, the accumulation of errors introduced by the optical flow methods causes color bleeding that impairs the visual effect. Recent DM-based methods~\cite{liu2023video,li2024towards,bozic2024versatile} apply cross-frame attention mechanisms to the image colorization backbone to facilitate multi-frame colorization. Nevertheless, the image colorization backbone is not inherently capable of time-domain understanding, and cross-frame attention only refer to a limited number of adjacent frames.

\subsection{Video Diffusion Models}
With the significant success of diffusion models\cite{rombach2022high} in the field of image generation, diffusion-based video generation frameworks \cite{guo_animatediff_2023,guo2025sparsectrl,blattmann2023videoldm,chen2023videocrafter1,blattmann_stable_nodate,chen2024videocrafter2,yang2024cogvideox,kong2024hunyuanvideosystematicframeworklarge} have emerged. There are two main technical routes for video generation based on diffusion models: (1) adding a temporal layer on the basis of image generation diffusion models, and (2) training large-scale video diffusion models using massive data. 

The representative work of Route 1 is AnimateDiff proposed by Guo et al. \cite{guo_animatediff_2023}. AnimateDiff extends the LDMs by incorporating a domain adapter, temporal transformer, and motion LoRAs\cite{hu2021lora}, thereby adapting it for video generation tasks. Based on this flexibile architecture, motion-driven video generation has developed rapidly \cite{hu2023animateanyone,wei2024dreamvideo,zhu2025champ,tian2025emo}, and a large number of works based on motion modeling have also emerged \cite{wang2024motionctrl,shi2024motion,zhao2025motiondirector}.  However, although two motion-related modules were designed and video training was incorporated, the generation capability is constrained by the pre-trained image generation model, resulting in suboptimal dynamic generation, especially in long video scenarios. 

The representative work of Route 2 is the large-scale trained video generation model. Stable Video Diffusion (SVD) \cite{blattmann_stable_nodate}, an open-source model with image-to-video generation capabilities, can generate 14 and 25 frame videos, providing high-quality video diffusion priors for downstream task research\cite{mimicmotion2024,huang2024lvcd}. DynamiCrafter\cite{xing2023dynamicrafter} offers both text-to-video and image-to-video generation capabilities and provides a detailed analysis of the text-image control conditions.

\section{Method}
Our goal is to design a pipeline for automatic colorization of grayscale video sequence $X^{gray}$ that can generate temporally consistent and color-saturated color sequences $\hat{Y}^{color}$. Firstly, in Section \ref{sec:svd}, we discuss the model architecture, where we re-purpose the video generation model SVD\cite{blattmann_stable_nodate} for the colorization task, enabling the model to generate videos guided by grayscale images. Once the model can achieve grayscale-controlled video generation, we introduce a global palette guidance to further refine the color control, supporting various flexible palette generation methods as described in Section \ref{sec:palette}.


\subsection{Video Diffusion Model for Video Colorization}\label{sec:svd}
Video diffusion models trained on large-scale image and video datasets exhibit excellent video quality and temporal consistency. We adopt the image-to-video model SVD as our backbone model. This choice is motivated by two aspects: (1) image-to-video models satisfy the requirements for grayscale fidelity in colorization tasks; (2) SVD is equipped with temporal modules in both the latent space and VAE, demonstrating stable performance across various downstream tasks. The framework is shown in Figure \ref{fig:overview}.

Specifically, during the training phase, we encode the grayscale input $X^{gray}$ and ground truth color sequence $Y^{color}$ to get the latent sequence $Z^{gray}$ and $Z^{color}$.  Subsequently, $Z^{color}$ is subjected to the forward noise process to obtain the noisy latent code $Z_t$:
\begin{equation}
    q(Z_t|Z^{color})\sim \mathcal{N}(Z_t; \sqrt{\Bar{\alpha_t}}Z^{color}, (1-\Bar{\alpha_t})\mathbf{I})
\end{equation}
where $\Bar{\alpha_t}=\Pi_{i=1}^t\alpha_i$ and $\alpha_i = 1-\beta_i$. $\beta_i$ represents the variance of the Gaussian noise added at the $i$-th diffusion step.
The grayscale latent code $Z^{gray}$ and $Z_t$ are concatenated along the channel dimension to get the input $Z_{in}$ for the denoising network:
\begin{equation}
    Z_{in} = \{\mathtt{concat}(z^i_{gray} ,z^i_t)\},\quad i = \{1,\cdots,N\}
\end{equation}
where $N$ denotes the input video length and $t$ denotes the denoising timestep.


During the inference process, the input $Z_{in}$ to the denoising U-Net consists of the encoded grayscale latent code $Z^{gray}$  and the sampled Gaussian noise $Z_T$. Inspired by MimicMotion\cite{mimicmotion2024}, we adopt the progressive approach for generating long videos with high temporal continuity. Specifically, we split the gray video sequence into segments with fixed length. Then, each segment is processed by the denoising U-Net. Finally, we obtain the color latent code by averaging the overlapped latent codes before sending it to the decoder.

\subsection{Palette Guidance}\label{sec:palette}
Image-to-video models face two key challenges in video colorization. First, without explicit color guidance, they tend to produce conservative results with muted, unsaturated colors. Second, since these models typically process only a limited number of frames at a time, maintaining consistent colors throughout the video becomes difficult.

Based on the main challenges of automatic video colorization, color guidance should consider both color saturation and cross-frame consistency. This requires color guidance to encourage saturated colors while maintaining the overall color style of the video, minimizing abrupt changes between frames. 

As shown in Figure\ref{fig:overview}, we design a global palette guide. The palette is composed of five distinct colors. Specifically, during the training process, we use the K-means clustering algorithm to categorize the colors within the reference frames into five palette colors. Subsequently, the palette vector $C_{palette}\in \mathbb{R}^{1\times15}$ constituted by these five colors is transformed through a straightforward color network composed of linear layers to obtain a color embedding $C_{emb}\in \mathbb{R}^{h\times w \times 320}$:
\begin{equation}
    C_{emb} = W_{proj} C_{palette}
\end{equation}
$C_{emb}$ is aligned with the first-layer features $F^{(1)}$ of the denoising U-Net in the channel dimension. $C_{emb}$ is then spatially broadcasted and fused with the first encoder layer:
\begin{equation}
  \mathbf{F}^{(1)} \leftarrow \mathbf{F}^{(1)} + \Phi(\mathbf{C}_{emb})
\end{equation}
Where $\Phi$ denotes spatial replication of the color embedding across $H\times W$ positions. This integration ensures that color information is consistently incorporated into the feature representation of each sequence, thereby enhancing the performance of the denoising process by providing additional contextual cues related to palette.

During training, we fine-tune the denoising U-Net while keep the parameters of VAE and CLIP encoder frozen. The overall objective function is defined in a similar way to stable video diffusion:
\begin{equation}
    \mathcal{L}=\mathbb{E}_{\epsilon\sim \mathcal{N}(0,1),t,z_t,I_{emb},C_{emb}}[||\epsilon-\epsilon_\theta(z_t;t,I_{emb},C_{emb})||_2^2]
\end{equation}
where $\epsilon$ is noise sampled from standard Gaussian. $t$ denotes denoising timestep. $I_{emb}$ represents the image embedding obtained by processing a randomly selected reference frame from the video through the CLIP image encoder\cite{radford2021learning}.
\begin{table*}[!hpt]
    \centering
    \caption{Performance comparison on DAVIS2017 test dataset. Bold and underlined characters represent the first and second rankings in the metric, respectively.}
    \label{tab:base}
    \vspace{4px}
    \begin{tabular}{l ccc ccc}
    \hline
    \textbf{Method}&  \textbf{Colorful$\uparrow$} & \textbf{FID$\downarrow$} & \textbf{FVD$\downarrow$} & \textbf{PSNR$\uparrow$}&\textbf{SSIM$\uparrow$} &\textbf{LPIPS$\downarrow$}\\
    \hline
    Deoldify\cite{antic_deoldify_nodate} & 20.47 & \underline{54.28} & \underline{648} & 23.57 & 0.9937 & 0.1972 \\ 
    VCGAN\cite{zhou2023vcgan} & 14.46 & 74.52 & 889 & 22.85 & 0.9053 & 0.3090 \\ 
    TCVC\cite{Liu2024Temporally} & \underline{21.77} & 71.25 & 776 & \textbf{24.69} & \underline{0.9973} & \underline{0.1798} \\  
    \textbf{Ours} & \textbf{22.64} & \textbf{53.76} & \textbf{590} & \underline{23.89} & \textbf{0.9993} & \textbf{0.1776} \\ 
    \hline
    \end{tabular}
    \end{table*}


\subsection{Flexible Palette Generation}

During the inference process, we provide various flexible methods for generating palette $C_{palette}$. For example, (a) by extracting from a colored reference image, (b) by randomly sampling the color space, or (c) by specifying colors provided by the user. Figure \ref{fig:overview} illustrates 3 methods to get the palette. 

Method (a) involves using K-means clustering on the provided colored reference image, following the same procedure as during training. The reference image can be any image or can be selected from frames processed by other colorization algorithms. We only extract the dominant colors from the reference image, and there are no strict requirements for the alignment between the reference image and the content of the video frames. In the practical application of colorizing old films, users can use images that reflect styles similar to the era of the current film as reference images for extracting color palettes. In this way, the model can adaptively align with the desired style.



Method (b), illustrated within the blue box, involves obtaining the palette through sampling from a trained GMM. We leverage the training data to fit the GMM. Specifically, we randomly select 10 frames from each video in the training set. To avoid a preference of conservative colors, we remove pixels with an RGB value variance less than 50. We then fit the GMM using the filtered pixels. During inference, a palette can be obtained by randomly sampling colors from the GMM. If the user has no specific preferences, sampling the color palette in this way can facilitate rapid automatic colorization. 

Method (c) involves the user directly specifying the colors of the palette. To reduce the user's workload, we propose the automated method illustrated in the green box. First, we use a tagging model\cite{zhang2023recognize} to extract content tags from the grayscale frame. Then, we employ a large language model\cite{achiam2023gpt} to generate colors corresponding to the extracted content. This approach can respond to objects within the video, offering more flexible and diverse color choices compared to sampling from a GMM. 

A comprehensive analysis of various color palette generation approaches and their visual impacts will be presented in Section \ref{sec: discuss}.



\begin{figure*}[!hpt]
\centering
\includegraphics[width=\textwidth]{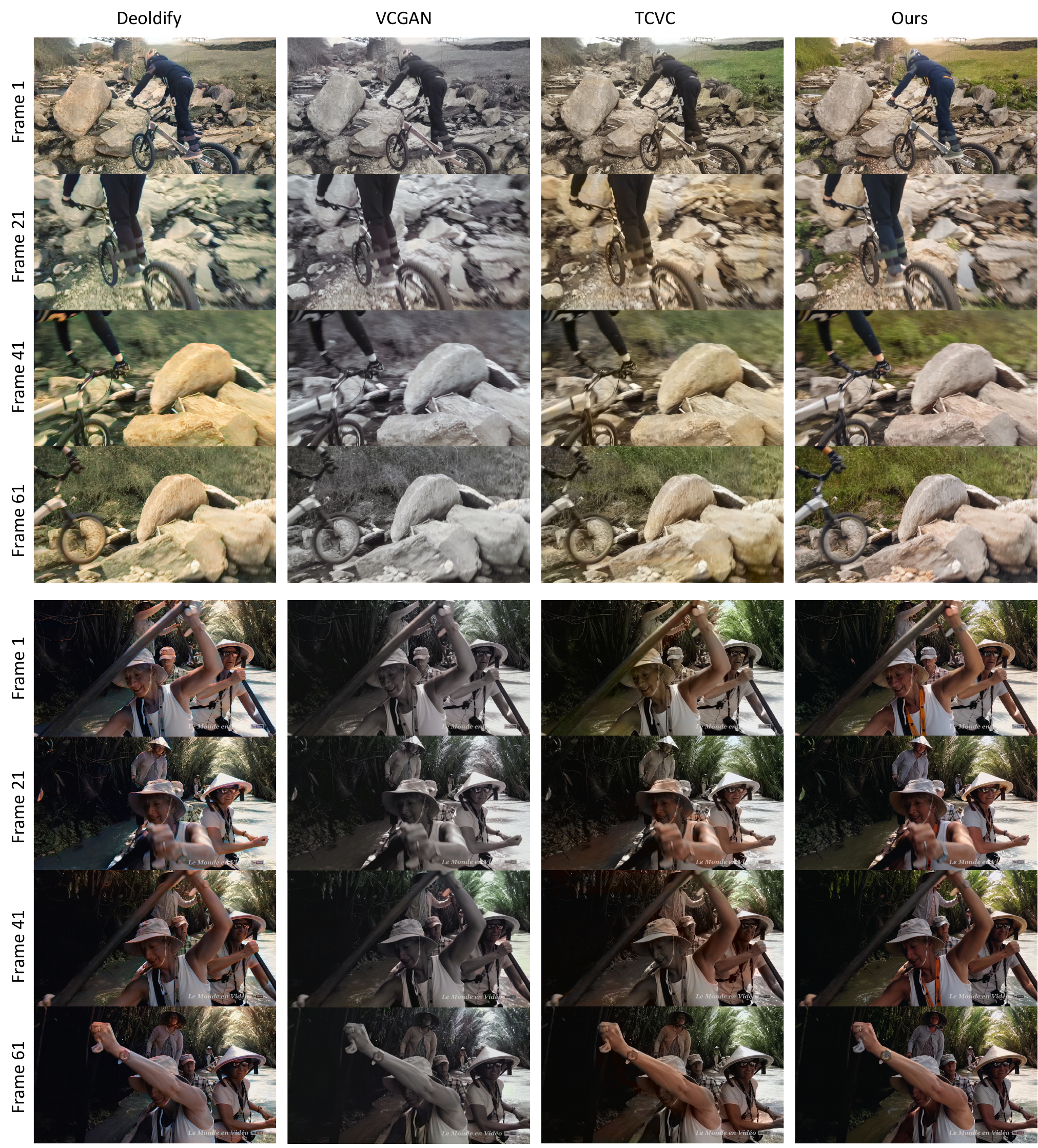}
\caption{Visual comparison of colorization results for 2 videos. From left to right, the colorization results for Deoldify\cite{antic_deoldify_nodate}, VCGAN\cite{zhou2023vcgan}, TCVC\cite{Liu2024Temporally}, and ours.}
\label{fig:base}
\end{figure*}

\section{Experiments}

We conduct three sets of experiments to evaluate our method: (1) comparison with state-of-the-art methods, (2) ablation studies, and (3) analysis of the palette-guidance capability.

\subsection{Implementation}
We leverage DAVIS2017\cite{ponttuset20182017davischallengevideo} benchmark to train and evaluate the proposed method. The training set contains 90 videos with an average length of 69 frames. During training, we resize the input video sequences to $1024\times 576$ and set the learning rate to $10^{-5}$ with a linear warmup for the first 500 iterations. The pre-trained weights are from the stable video diffusion 1.1 image-to-video model.

We train the denoising U-Net and the linear color network while keeping the VAE and CLIP image encoder frozen. 


\subsection{Comparison with State-of-the-arts}
We conduct qualitative comparison experiments with existing automantic colorization methods, including Deoldify\cite{antic_deoldify_nodate} which is modified from an image colorization method, VCGAN\cite{zhou2023vcgan} and TCVC\cite{Liu2024Temporally}. The inference for the baseline methods employs the model weights and inference scripts released officially. The \textbf{visual comparison} is shown in Figure\ref{fig:base}. We select frames with a long temporal span for comparison to highlight the temporal consistency of the models. The frames from top to bottom correspond to the 1st, 21st, 41st, and 61st frames of the video.

In the simple scene (see the first video in Figure \ref{fig:base}), Deoldify's results exhibit monotonous color in each individual frame, with insufficient contrast and saturation. Moreover, there is a significant color tone difference between frames that are farther apart, leading to a poor overall visual effect in the video. VCGAN shows extremely low color saturation, which can be considered a failure in colorization for this particular case. TCVC performs relatively well but exhibits semantic errors in the second frame, where the person is not recognized and is assigned the same color as the surrounding rocks, indicating a lack of temporal stability. Our method provides stable, saturated, and semantically reasonable results in this case, with a natural overall visual effect. 

In complex scene (see the second video in Figure \ref{fig:base}), all methods are capable of coloring the green plants in the background. However, when it comes to coloring the foreground figures, Deoldify exhibits color bleeding at the junctions of different objects and fails to handle details such as the pendant on the woman's neck. VCGAN once again fails in the colorization task. TCVC, which outperforms Deoldify in simple scenes, encounters severe semantic errors in complex scenes. In the first frame, the skin tone of the foreground figure is incorrectly assigned green, although this issue is somewhat mitigated in subsequent frames, the colors of the background figures remain unsatisfactory. Moreover, due to color bleeding from the surrounding environment and the strange skin tone, the overall visual effect is poor. Additionally, TCVC suffers from the same problem as Deoldify in missing the color details of the pendant. These issues in baseline methods primarily stem from their reliance on GAN-based backbones, which struggle to generate diverse colors and fine-grained details. Furthermore, their single-frame processing approach lacks the capability for high-quality long-range perception, resulting in poor color consistency across frames. 

Our method, on the contrary, provides vivid and reasonable colorization for both the foreground and background, with accurate handling of details. There is no noticeable color inconsistency between frames that are far apart, and the overall visual effect surpasses the compared methods. This superiority is attributed to our multi-frame processing framework, combined with a consistent global guidance mechanism, ensures high-quality long-range color consistency and temporal stability, addressing the limitations of baseline methods. Additionally, our colorization results do not rely on the ground truth color palette; the color palette used here is obtained through random sampling from the trained GMM.

To provide a comprehensive comparison between the proposed method and the baseline methods, we conduct \textbf{quantitative experiments}. We select the metrics widely used in recent years to evaluate image coloring tasks, the Colorful metric \cite{hasler2003measuring} and the Fréchet Inception Distance (FID) \cite{heusel2017gans}, to evaluate the color saturation and image quality of each frame, respectively. We also introduce the Fréchet Video Distance (FVD) \cite{unterthiner2019fvd} to evaluate the video quality. Moreover, we also compare the commonly used metrics PSNR, SSIM and LPIPS\cite{zhang2018unreasonable}, to provide a comprehensive evaluation.

Table\ref{tab:base} presents the results of our quantitative experiments. Our method demonstrates significant advantages, achieving the best quantitative results in all metrics except PSNR, where it is second only to TCVC. This aligns with the qualitative analysis in the previous section, indicating that our video colorization algorithm can provide semantically reasonable, saturated, and temporally stable color videos. The improvement in realism can be attributed to the introduction of an advanced diffusion model as the backbone, which enhances the generation of high-fidelity details. The enhancement in color saturation is primarily due to our innovative palette guidance mechanism, which ensures vibrant and visually appealing colorization. Furthermore, the overall video quality improvement stems from two key factors: (1) the integration of a video generation model for multi-frame processing, which ensures temporal consistency across frames, and (2) the consistent global guidance provided by our palette mechanism, which maintains coherence throughout the entire video sequence.
\begin{figure*}[ht]
\centering
\includegraphics[width=\textwidth]{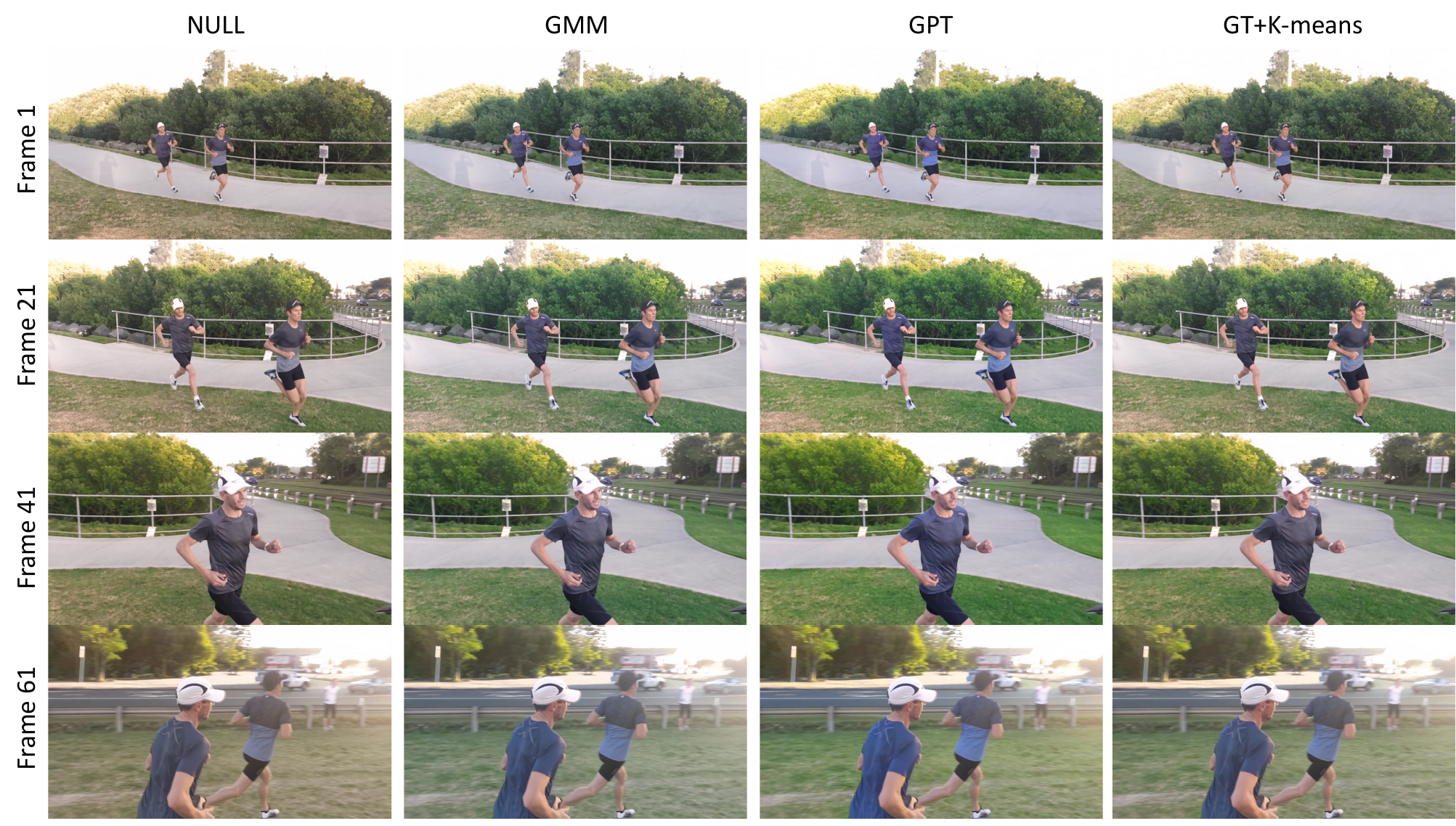}
\caption{Visual effects of ablation studies. The first column (NULL) shows the colorization results without palette guidance. The other columns display the colorization results of model variants with different palette generation methods.}
\label{fig:ab}
\end{figure*}
\subsection{Ablation Studies}\label{sec:ab}
This section systematically evaluates two core innovations of our framework: (1) the necessity of global palette guidance and (2) the impact of different palette generation methods. We design four experimental configurations:
\begin{enumerate}
    \item[$\mathcal{A}$]: Without palette guidance (retrained model)
    \item[$\mathcal{B}$]: Random palette from trained GMM distribution
    \item[$\mathcal{C}$]: LLM-assisted palette (GPT-4o as generator)
    \item[$\mathcal{D}$]: Ground-truth first-frame extracted palette
\end{enumerate}

All experiments maintain identical training configurations except for palette inputs. For fair comparison in Experiment $\mathcal{A}$, we completely remove the palette branch and retrain the model. Experiments $\mathcal{B-D}$ preserve the original architecture but vary in palette sources: $\mathcal{B}$ samples colors from our trained Gaussian mixture model, $\mathcal{C}$ leverages GPT for semantic-aware color selection, while $\mathcal{D}$ adopts an oracle strategy using ground truth references.

\begin{table}[t]
    \centering
    \caption{Quantitative results of ablation study on global palette guidance}
    \label{tab:wp2-ab-1}
    \vspace{4px}
    \begin{tabular}{cc ccc}
    \hline
    \textbf{Exp.}&\textbf{Palette}&  \textbf{Colorful$\uparrow$} & \textbf{FID$\downarrow$} & \textbf{FVD$\downarrow$} \\ \hline
    $\mathcal{A}$ &NULL& 19.34 & 54.63 & 628 \\ 
    $\mathcal{B}$&GMM&22.64 & 53.76 & \underline{590}\\ 
    $\mathcal{C}$ &GPT& \textbf{34.92}    & \underline{53.47} & 637 \\  
    $\mathcal{D}$ &GT+K-means& \underline{25.55}    & \textbf{50.40} & \textbf{562}\\  
    \hline
    \end{tabular}
    \end{table}

As quantified in Table \ref{tab:wp2-ab-1}, Experiment A demonstrates significant performance degradation across key metrics:

\noindent\textbf{Colorfulness}: $30.2\%$ lower than average of $\mathcal{B-D}$ ($\Delta = 8.36$).

\noindent\textbf{Visual Quality (FID)}: $2.1$ worse than average of $\mathcal{B-D}$.

\noindent\textbf{Temporal Consistency (FVD)}: $31.7$ worse than average of $\mathcal{B-D}$.

This empirically confirms our hypothesis that palette guidance resolves ambiguity in ill-posed colorization tasks. The visual comparison in Figure \ref{fig:ab} further reveals that $\mathcal{A}$ produces unsaturated results. In contrast, $\mathcal{B-D}$ equipped with the palette guidance achieves saturated colors and better visual effects.

Our analysis of palette generation methods reveals distinct performance characteristics across key metrics: GMM-sampled palettes exhibit slightly lower color saturation (Colorfulness) compared to the other two approaches, while GPT-generated palettes achieve the highest color vibrancy. The GT-extracted palette demonstrates superior performance in both image quality (FID) and temporal consistency (FVD), indicating its effectiveness in maintaining photorealistic appearance and coherent video transitions. This result demonstrates that while GMM-based methods achieve better temporal consistency in automatic colorization tasks (as reflected in FVD metrics), LLM-driven approaches excel at creative color enhancement (evidenced by Colorfulness scores), whereas introducing color frame extraction for palette generation yields superior visual quality in both image (FID) and video (FVD) domains.

\subsection{Flexibility of palette-guidance}\label{sec: discuss}
\begin{figure}[t]
\centering
\includegraphics[width=0.9\linewidth]{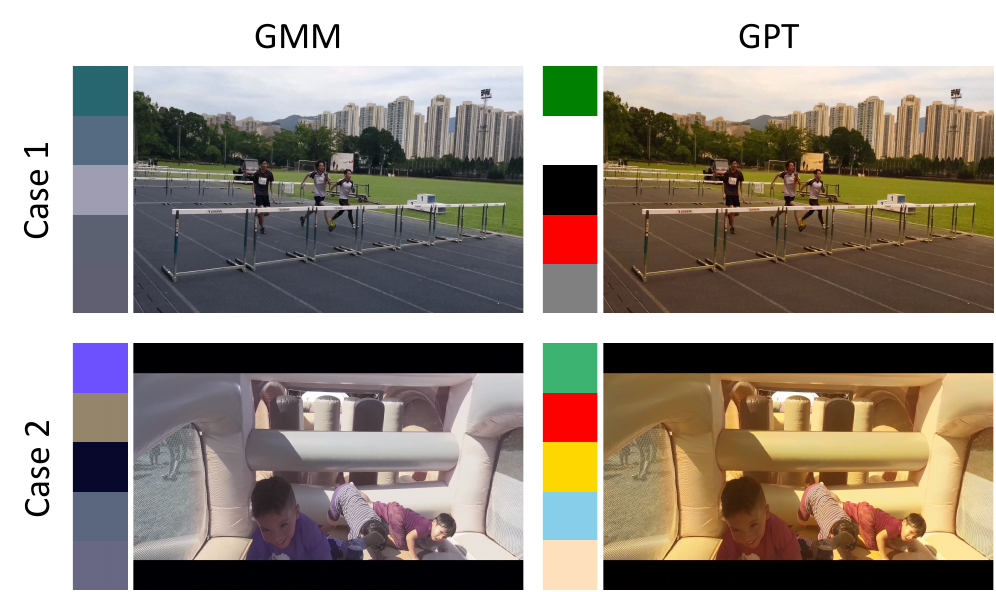}
\caption{Diverse colorization results with distinct palettes.}
\label{fig:palette}
\end{figure}

We discuss the flexibility of palette guidance. 
Given that the colorization task is ill-posed, the color selection for many objects, particularly man-made ones, is not unique. There are various reasonable colors for the same image. We illustrate the colorization results corresponding to different palettes in Figure \ref{fig:palette}. We compare two different automatic palette generation methods, including random sampling from the trained GMM and GPT assisted generation, together with their corresponding colorization results. These two methods can automatically generate palettes without relying on external prompts. Focusing on man-made objects, the colors of the track in the left image and the toys as well as the child's clothing in the right image are significantly influenced by the palette. Even with substantial differences in palette colors, the generated images remain semantically reasonable. These cases demonstrate that our palette can achieve diverse and controllable colorization.


\section{limitation}
While our method achieves effective global color style control by adaptively applying palette colors to video content, it currently lacks precise instance-level color manipulation capabilities. The adaptive palette propagation mechanism, though robust for holistic style transfer, cannot isolate and control colors for specific objects or regions (e.g., recoloring individual vehicles in traffic scenes or modifying clothing colors in human-centric videos). A promising extension of this work would involve integrating spatiotemporal object tracking and semantic-aware hint color guidance with our framework. 
\section{Conclusion}
In this paper, we propose an automatic video colorization method via palette guidance. To address the challenge of color inconsistency in video colorization, we introduce the image-to-video generation algorithm, Stable Video Diffusion, into the video colorization task. To tackle the challenge of color unsaturation, we design a global palette control. Our palette significantly enhances color saturation while maintaining video color consistency. Furthermore, our palette supports flexible generation methods, including automatic generation, user specification, and reference image extraction modes. Both quantitative and qualitative experimental results demonstrate that our proposed method can achieve highly saturated, high-quality, and consistent video colorization.

\bibliographystyle{IEEEbib}
\bibliography{ref}

\end{document}